\documentclass{article}

\usepackage{arxiv}

\usepackage[utf8]{inputenc} 
\usepackage[T1]{fontenc}    
\usepackage{hyperref}
\usepackage{url}            
\usepackage{booktabs}       
\usepackage{amsfonts}       
\usepackage{nicefrac}       
\usepackage{microtype}      
\usepackage{graphicx}
\usepackage[numbers,sort&compress]{natbib}
\usepackage{doi}
\usepackage[version=4]{mhchem}
\usepackage{siunitx}
\DeclareSIUnit\Molar{M}
\usepackage{bm}

\title{Improving Neuropathological Reconstruction Fidelity via AI Slice Imputation}

\author{
Marina Crespo Aguirre$^{1,2}$ \and 
Jonathan Williams-Ramirez$^{1}$ \and 
Dina Zemlyanker$^{1}$ \and 
Xiaoling Hu$^{1}$ \and
Lucas J. Deden-Binder$^{1}$ \and 
Rogeny Herisse$^{1}$ \and 
Mark Montine$^{3}$ \and 
Theresa R. Connors$^{4}$ \and
Christopher Mount$^{5}$ \and 
Christine L. MacDonald$^{6}$ \and 
C. Dirk Keene$^{3}$ \and 
Caitlin S. Latimer$^{3}$ \and 
Derek H. Oakley$^{4}$ \and 
Bradley T. Hyman$^{4}$ \and 
Ana Lawry Aguila$^{1}$\thanks{These authors contributed equally to this work.} \and
Juan Eugenio Iglesias$^{1,7,8}$\footnotemark[1]
\thanks{Corresponding author: \texttt{jiglesiasgonzalez@mgh.harvard.edu}}
\\
\\
$^{1}$ Martinos Center for Biomedical Imaging, Massachusetts General Hospital and Harvard Medical School, United States \\
$^{2}$ Federal Institute of Technology (ETH), Zurich, Switzerland \\
$^{3}$ BioRepository and Integrated Neuropathology (BRaIN) Laboratory and Precision Neuropathology Core,\\
\quad University of Washington School of Medicine, Seattle, United States \\
$^{4}$ Massachusetts Alzheimer Disease Research Center, MGH and Harvard Medical School, Charlestown, United States \\
$^{5}$ Department of Pathology, Massachusetts General Hospital and Harvard Medical School, Boston, MA, United States \\
$^{6}$ Department of Neurological Surgery, University of Washington School of Medicine, Seattle, United States \\
$^{7}$ Hawkes Institute, University College London, United Kingdom \\
$^{8}$ Computer Science and Artificial Intelligence Laboratory, Massachusetts Institute of Technology, United States
}

\begin{document}

\maketitle

\begin{abstract}
Neuropathological analyses benefit from spatially precise volumetric reconstructions that enhance anatomical delineation and improve morphometric accuracy. Our prior work has shown the feasibility of reconstructing 3D brain volumes from 2D dissection photographs. However these outputs sometimes exhibit coarse, overly smooth reconstructions of structures, especially under high anisotropy (i.e., reconstructions from thick slabs). Here, we introduce a computationally efficient super-resolution step that imputes slices to generate anatomically consistent isotropic volumes from anisotropic 3D reconstructions of dissection photographs. By training on domain-randomized synthetic data, we ensure that our method generalizes across dissection protocols and remains robust to large slab thicknesses. The imputed volumes yield improved automated segmentations, achieving higher Dice scores, particularly in cortical and white matter regions. Validation on surface reconstruction and atlas registration tasks demonstrates more accurate cortical surfaces and MRI registration. By enhancing the resolution and anatomical fidelity of photograph-based reconstructions, our approach strengthens the bridge between neuropathology and neuroimaging. Our method is publicly available at \url{https://surfer.nmr.mgh.harvard.edu/fswiki/mri_3d_photo_recon}.
\end{abstract}


\section{Introduction}

\begin{figure}[h!]
    \centering
    \includegraphics[width=\linewidth]{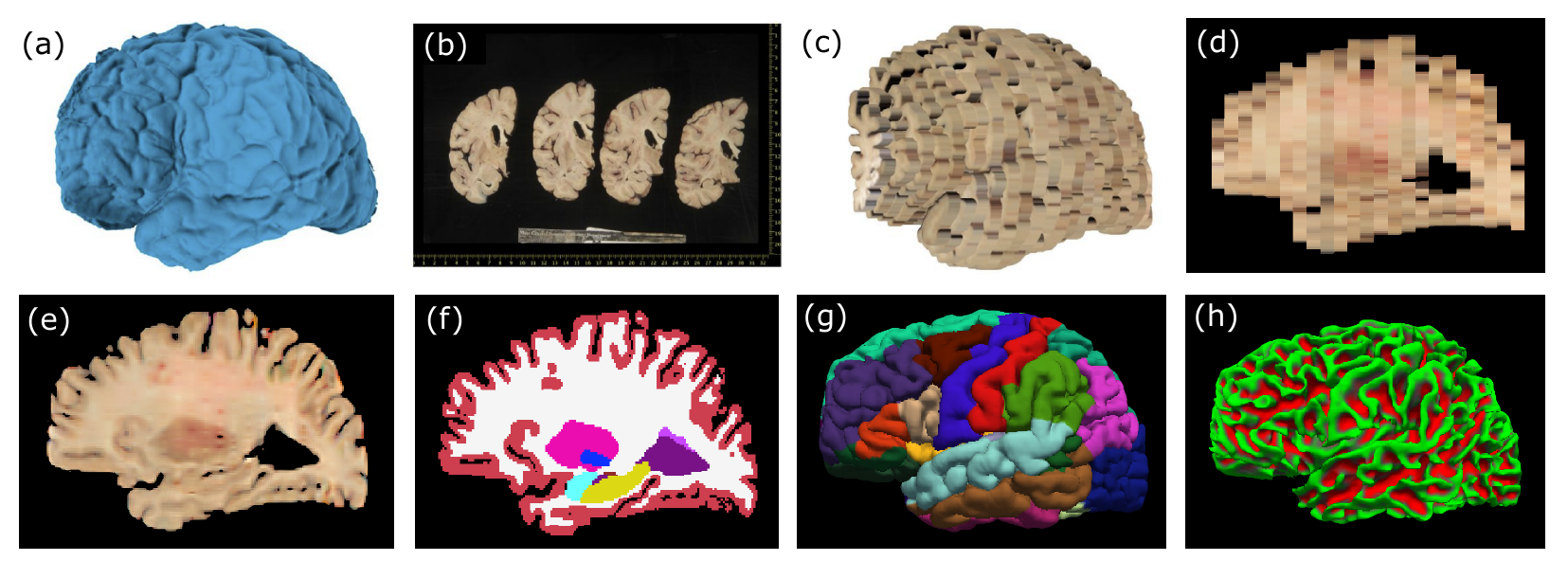}
    \caption{Input and outputs for a sample case. (a)~3D Surface scan of the left human hemisphere acquired prior to dissection. (b)~Photographs of dissected coronal slabs (thickness$\approx$8~mm), post pixel size calibration, with digital ruler overlaid. (c)~3D reconstruction of the photographs into an imaging volume. (d)~Sagittal cross-section of the volume prior to imputation. (e)~Sagittal cross-section of the volume after imputation with our approach, which recovers high-resolution detail. (f)~Corresponding slice of automated segmentation obtained from (e) with Photo-SynthSeg \citep{gazula2024machine}. (g) Pial surface with overlaid parcellation, obtained from (e) using Recon-Any \citep{gopinath2025recon}. (h) White matter surface obtained with Recon-Any.}
    \label{fig:motivation_intro}
\end{figure}

We have recently shown \citep{gazula2024machine} that vast archives of postmortem brain dissection photographs (routinely acquired by brain banks for documentation) can be repurposed for quantitative 3D analysis. This method relies on turning 2D dissection photographs into anatomically faithful 3D brain volumes, using an external reference such as a brain atlas or a surface scan acquired prior to slicing \citep{salvi2004pattern}. These 3D reconstructions enable tasks such as atlas registration, segmentation, or volumetry without the need for an \emph{ex vivo} MRI -- a technique that requires highly specific expertise and substantial resources. As illustrated in Figure~\ref{fig:motivation_intro}a–d, our framework converts simple archival photographs into spatially consistent volumetric representations, providing a cost-effective bridge between neuropathology and neuroimaging.

Despite this progress, a key limitation remains: dissected slabs are typically several millimeters thick, leaving potentially large gaps in between. This may hinder the precision of downstream morphometric analyses -- particularly in the cortex, where accurate surface geometry is essential. Even algorithms designed to tolerate irregular slice spacing, such as our previous reconstruction pipeline \citep{gazula2024machine} or the domain-agnostic segmentation framework SynthSeg \citep{billot2023synthseg,billot2023robust}, cannot fully recover anatomically plausible details when faced with such anisotropy. Similar challenges have been reported in other modalities, where sparse-slice sampling degrades cortical and subcortical quantification \citep{chen2018efficient,pham2019multiscale,zhao2020smore}. Hence, addressing the through-plane resolution gap is essential for anatomically accurate 3D reconstruction.

Machine learning offers a natural solution to this problem through slice imputation via super-resolution \citep{xia2021recovering,wu2022slice,liu2025sagcnet}. These approaches aim to infer missing anatomical detail between sparsely sampled planes, such that these details are accurately imputed in the reconstructed volume. 
In the formulation proposed here, a 2D U-Net \citep{ronneberger2015u} receives two adjacent dissection slices as input and predicts an anatomically consistent ``missing'' slice at a specified coordinate between them. This enables flexible interpolation across variable slice spacings, by iterative application at any desired through-plane spacing (typically 1~mm) at test time. Furthermore, operating on one slice pair at a time offers several advantages: it simplifies memory requirements, avoids the need for fixed inter-slice distances, and enables efficient inference even on modest computational hardware.

Applying this approach to dissection data in a supervised manner would require paired high- and low-resolution training examples, thus necessitating a near-impossible acquisition of ultra-thin ($\sim$1 mm) slabs, followed by selective omission to create training pairs. Instead, a synthetic training paradigm provides a practical and generalizable alternative. By simulating arbitrary image contrast and slice spacings from existing 1 mm isotropic MRI datasets, one can generate large, domain-randomized training sets that encompass diverse geometries, contrasts, and tissue conditions. We have successfully used this strategy in prior work for domain-agnostic brain MRI analysis \citep{billot2023synthseg,iglesias2023synthsr,gopinath2025recon,gopinath2024synthetic}, where it enabled robust, cross-protocol generalization to real dissection photographs acquired under varying illumination, camera, and tissue fixation.

Building upon our prior framework \citep{gazula2024machine}, we introduce this imputation step within the 3D reconstruction pipeline. Trained entirely on synthetic data, our model generates anatomically consistent isotropic volumes from stacks of photographs of (potentially thick) coronal slabs, while remaining agnostic to slab spacing (thickness) and acquisition  (Figure~\ref{fig:motivation_intro}e). We demonstrate that our machine learning imputation method: \textit{(i)}~produces more anatomically accurate 3D reconstructions, substantially improving automated segmentations (Figure~\ref{fig:motivation_intro}f), \textit{(ii)}~produces smoother, more anatomically faithful surfaces, which in turn yield better cortical thickness estimates (Figure~\ref{fig:motivation_intro}g-h); and \textit{(iii)}~considerably improves atlas registration.


\section{Results}
\subsection{Qualitative Results}
We revisit two datasets from the previous publication, consisting of 3D reconstructions from 2D slabs of real specimens from the Alzheimer's Disease Center at the University of Washington (UW) and the Massachusetts Alzheimer's Disease Research Center (MADRC). 
The UW dataset is composed of homogeneous slices of 4 mm thickness; 8 mm and 12 mm variants are generated by subsampling the available photographs by a factor of 2 and 3, respectively.
The MADRC dataset comprises of reconstructions with heterogeneous slices of approximately 8 mm thickness.

\begin{figure}[h!]
    \centering
    \includegraphics[width=0.95\linewidth]{ 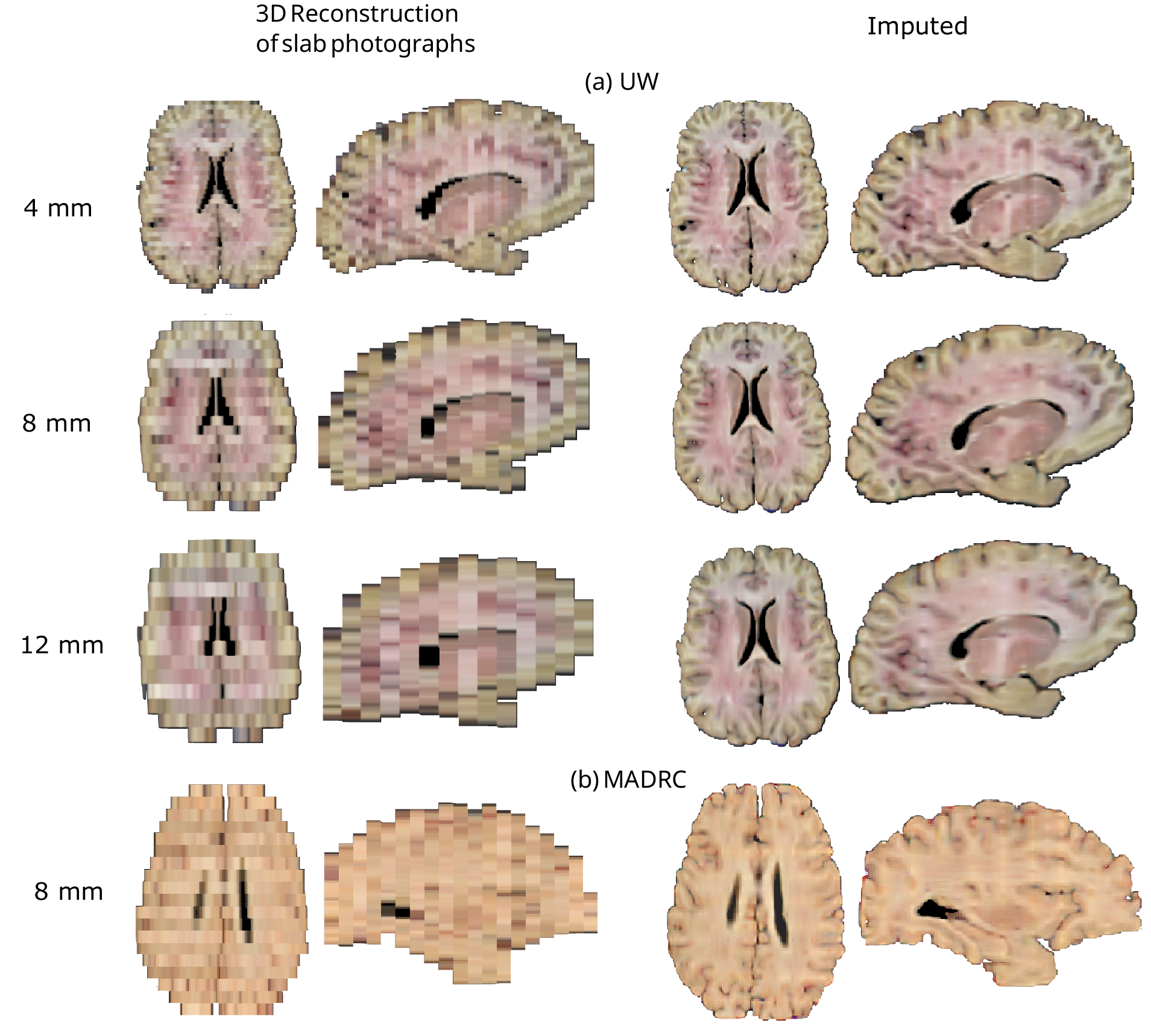}
    \caption{Axial and sagittal views of 3D reconstructions before and after imputation. (a)~Sample case
    from the UW dataset, showing the original 4-mm slabs and the 8-mm and 12-mm variants. (b) Sample case from the MADRC dataset, comprising 8-mm thick slices. On the left, 3D reconstructions from our previous work \citep{gazula2024machine}; 
    on the right, results of the proposed imputation.}
    \label{fig:task_0_reconstructions}
\end{figure}

Figure~\ref{fig:task_0_reconstructions} shows the original reconstructions from our previous work and the outputs of our new imputation method on two sample specimens from the UW (Panela)  and MADRC datasets (Panel~b).
At 4 mm thickness, larger brain regions can still be distinguished 
on the original 3D reconstructions of slab photographs (e.g.,  thalamus, ventricles).
However, as slab thickness increases, the 3D reconstructions appear coarser, with reduced definition of these anatomical regions. In contrast, the cortical and subcortical regions are better defined in the reconstructions generated by our imputation method (Figure ~\ref{fig:task_0_reconstructions}.b). The improvements from our method are particularly evident on  reconstructions from thicker slabs (8-12 mm), especially in the depiction of the cortical folding, which appears much less voxelated -- truly bringing the white matter and pial surfaces to life.

\subsection{Surface Reconstruction}

We evaluate the impact of our imputation method at a cortical level using the pipeline ``Recon-Any'' \citep{gopinath2025recon} distributed with FreeSurfer. This tool extracts cortical surface meshes of brain imaging volumes of any modality, contrast, and resolution, without retraining or fine-tuning. Along with the surface meshes, ``Recon-Any'' predicts 34 cortical parcellations, providing a complete labeling of cortical gyri and sulci.
Using this software, we extract the pial and white matter surfaces for the original 3D reconstructions of stacked photographs and the reconstructions generated by our imputation method. We assess the accuracy of these meshes by comparing them to cortical surfaces derived from 1 mm~isotropic FLAIR MRI scans acquired \textit{ex vivo} prior to dissection, which we use as gold standard.

\begin{figure}[h!]
    \centering
        \includegraphics[width=\linewidth]{ 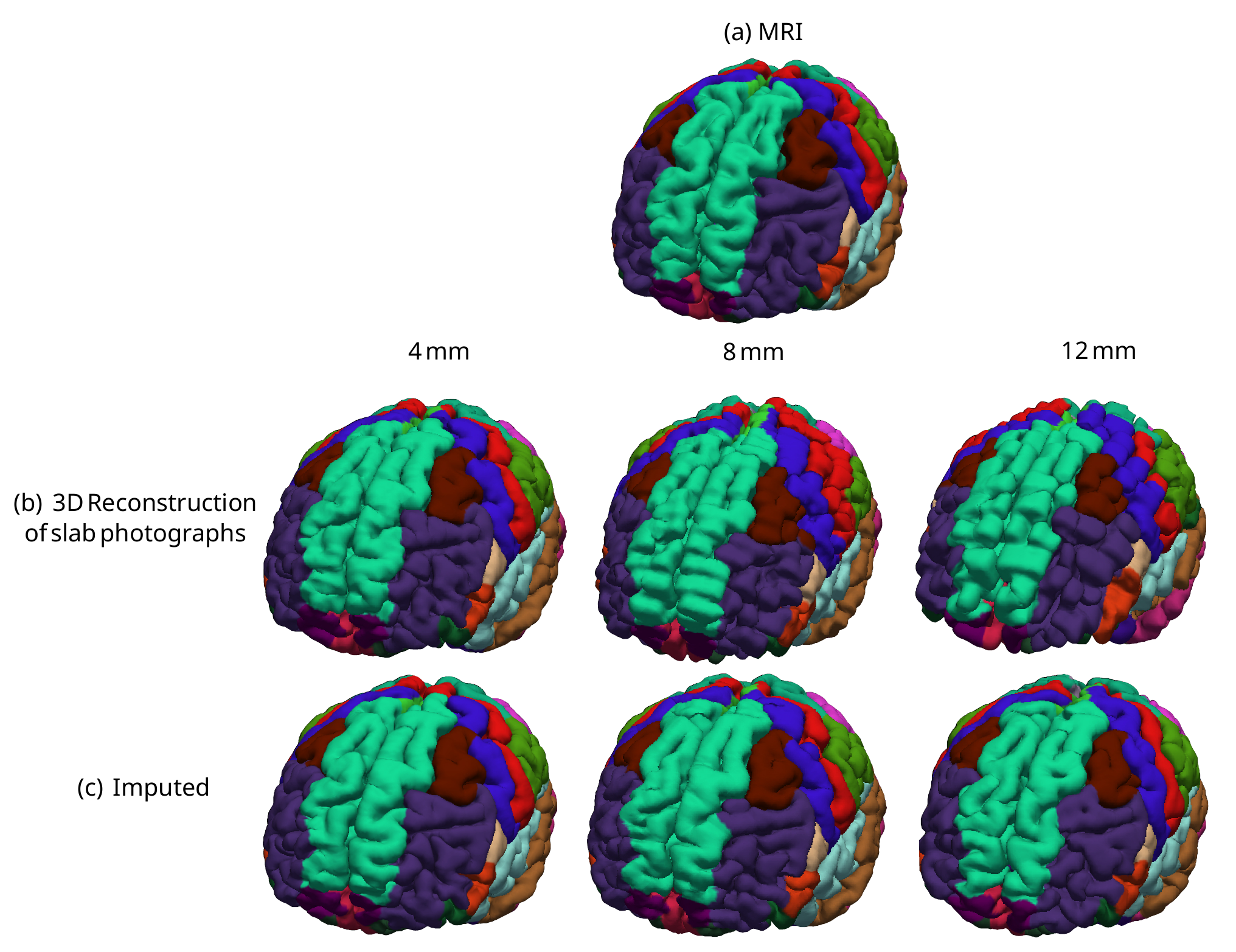}
    \caption{Pial surface meshes with overlaid parcellations of Recon-Any on one case from the UW dataset at three slab thicknesses, with and without imputation. (a)~Reference pial surface from the MRI (gold standard); (b)~surface results from the 3D reconstruction of slab photographs, and (c)~surface results from the proposed imputations.}
    \label{fig:task_1_uw_surface_recon_example}
\end{figure}

Figure~\ref{fig:task_1_uw_surface_recon_example} and supplementary Figure S~\ref{app:task_1_madrc_illustration_surfaces}, show the pial surfaces of two cases from the UW and MADRC dataset, respectively. 
Without imputation, the generated surfaces appear blocky and voxelated, especially in regions oriented perpendicular to the slicing direction. In contrast, when our proposed imputation method is applied, we obtain smoother and more anatomically plausible boundaries, even at the largest slab thickness (12 mm).

\begin{figure}[h!]
    \centering
    \includegraphics[width=\linewidth]{ 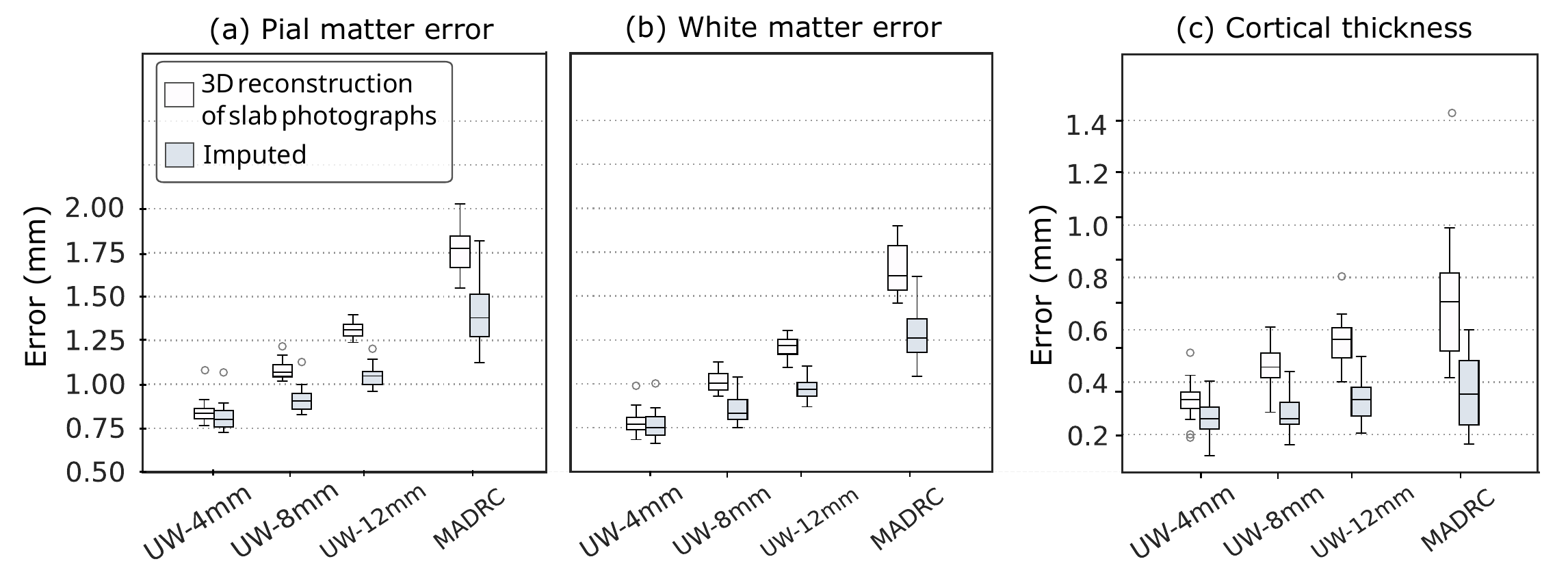}
    \caption{Box plots of surface errors for Recon-Any on 3D reconstructions of photographs, with and without imputations, for both the MADRC and UW datasets. (a)~Pial surface error (mm), (b)~white matter surface error (mm), and (c)~cortical thickness error (mm).  
    The decrease in error due to the imputation is significant at $p<0.001$ in all cases. }
    \label{fig:task_1_surferrors}
\end{figure}

For each brain region, we compute the closest-point distance between the cortical meshes of the photo reconstructions and the gold-standard meshes derived from the MRIs. We estimate the mean distance error within each parcel, and average across the 34 labeled cortical areas, yielding a single value of distance error for pial and white matter surfaces per specimen.
Similarly, we compute the cortical thickness error, measuring the point-distance between white matter and pial matter surface meshes. Thickness errors are computed by comparing cortical thicknesses of photo-reconstructions against the gold-standard thicknesses.
Box plots in Figure~\ref{fig:task_1_surferrors} report the surface and thickness errors for the UW and MADRC datasets. The surface visualizations of cortical thickness errors from both datasets are also shown in Figure S~\ref{app:task_1_uw_madr_cortical_thickness_ilustration} in the Supplementary Material. 

Although the error magnitudes grow with slab thickness, our imputation method consistently achieves lower errors compared to the volumes generated using the baseline approach.
We see statistically significant improvements for all metrics, datasets, and resolutions, with a decrease in the surface errors of up to approximately 0.4 mm, compared to the baseline reconstructions (Wilcoxon $p < 0.001$, Table S~\ref{app:surface_cortical_errors_updated}).  This improvement is particularly noticeable on thicker slabs (8 mm for both datasets, and 12 mm for the UW dataset). 

\subsection{Volume Segmentation} 
We explore the performance of our imputation method on an automated segmentation task.
Gold-standard segmentations  are obtained with ''SynthSeg''  \citep{billot2023synthseg} on the MRIs; the reconstructed imputations are segmented in the same way. SynthSeg is a 3D segmentation model capable of handling input scans of any contrast and resolution without requiring retraining or fine-tuning; it returns  high-resolution 1 mm isotropic segmentations independently of the resolution of the input scan.
The baseline photo-reconstructions, are instead segmented with ''Photo-SynthSeg'' \citep{gazula2024machine}, a version of ''SynthSeg'' trained explicitly to deal with large coronal spacings from 3D reconstructions of photographs. 
We compute region-specific Dice scores to quantify the agreement between the segmentation of the photo reconstruction (with and without imputation) and the gold standard.

\begin{figure}[h!]
    \centering
    \includegraphics[width=0.86\linewidth]{ 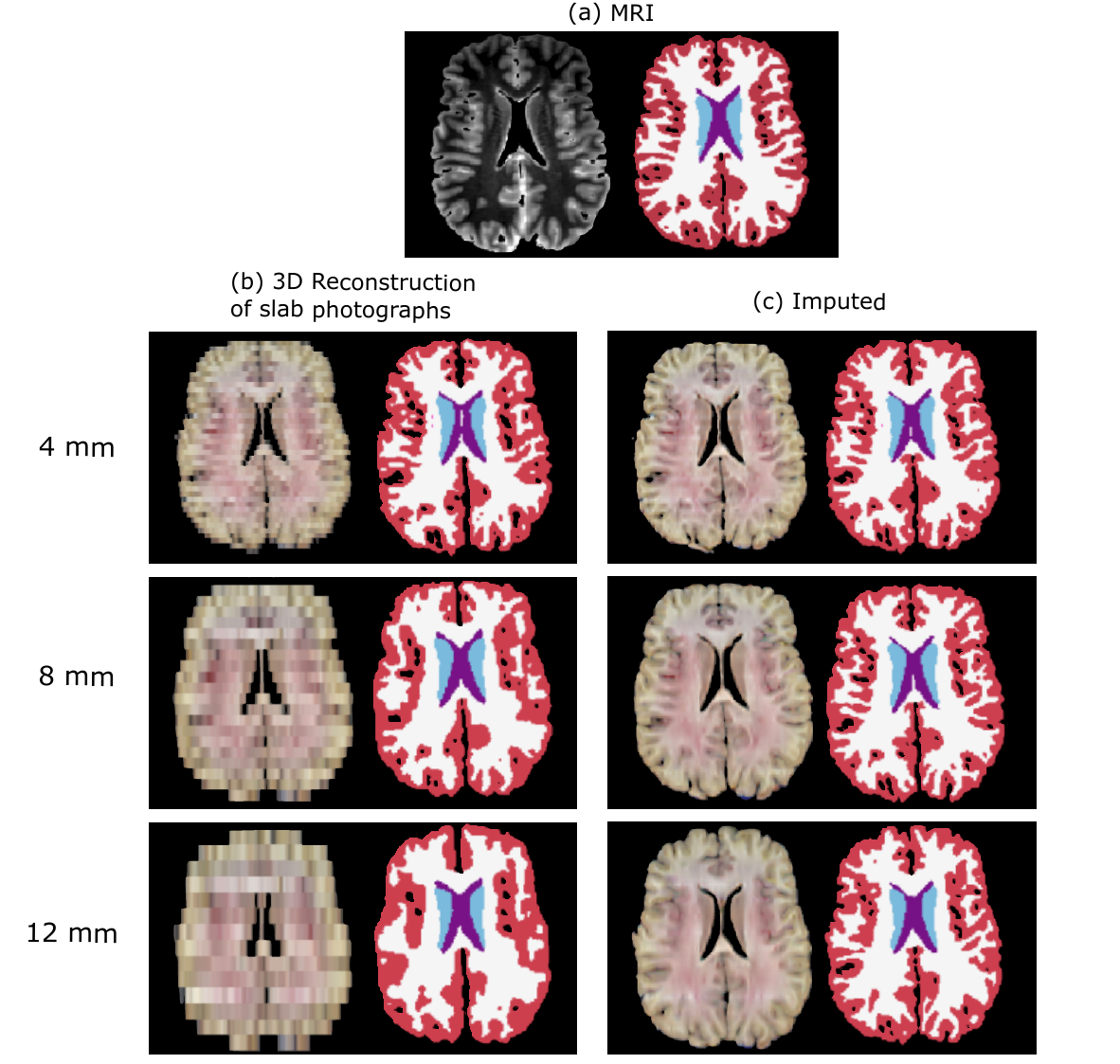}
    \caption{Axial views of automated segmentations on one example from the UW dataset at three slab thicknesses (4, 8, 12 mm). (a)~Gold-standard segmentation from the MRI. (b)~Segmentations on the 3D reconstructions from the original slab photographs. (c)~Segmentations from imputed reconstructions.}
    \label{fig:photosynthseg_segmentations_uw}
\end{figure}

Figure~\ref{fig:photosynthseg_segmentations_uw}, and Supplementary Figure S~\ref{app:photosynthseg_segmentations_madrc}, show the segmentation results for two sample cases from the UW and MADRC datasets, respectively.
Increasing the slab thickness degrades the segmentation performance on the baseline photo stacks, with the largest errors observed in the cortical and white matter regions, compared to the gold-standard segmentations. This is noticeable, e.g., in Figure~\ref{fig:photosynthseg_segmentations_uw}b (12~mm), where the segmentation map shows a blocky appearance, with a discontinuous cortical ribbon and holes along the sulcal banks and gyral crowns. 
A similar effect is observed in the segmentations from the MADRC example (Figure S~\ref{app:photosynthseg_segmentations_madrc}b). Our imputation method greatly improves the segmentation detail in both  datasets (Figure~\ref{fig:photosynthseg_segmentations_uw}c and Figure S~\ref{app:photosynthseg_segmentations_madrc}c).

\begin{figure}[h!]
    \centering
    \includegraphics[width=1\linewidth]{ 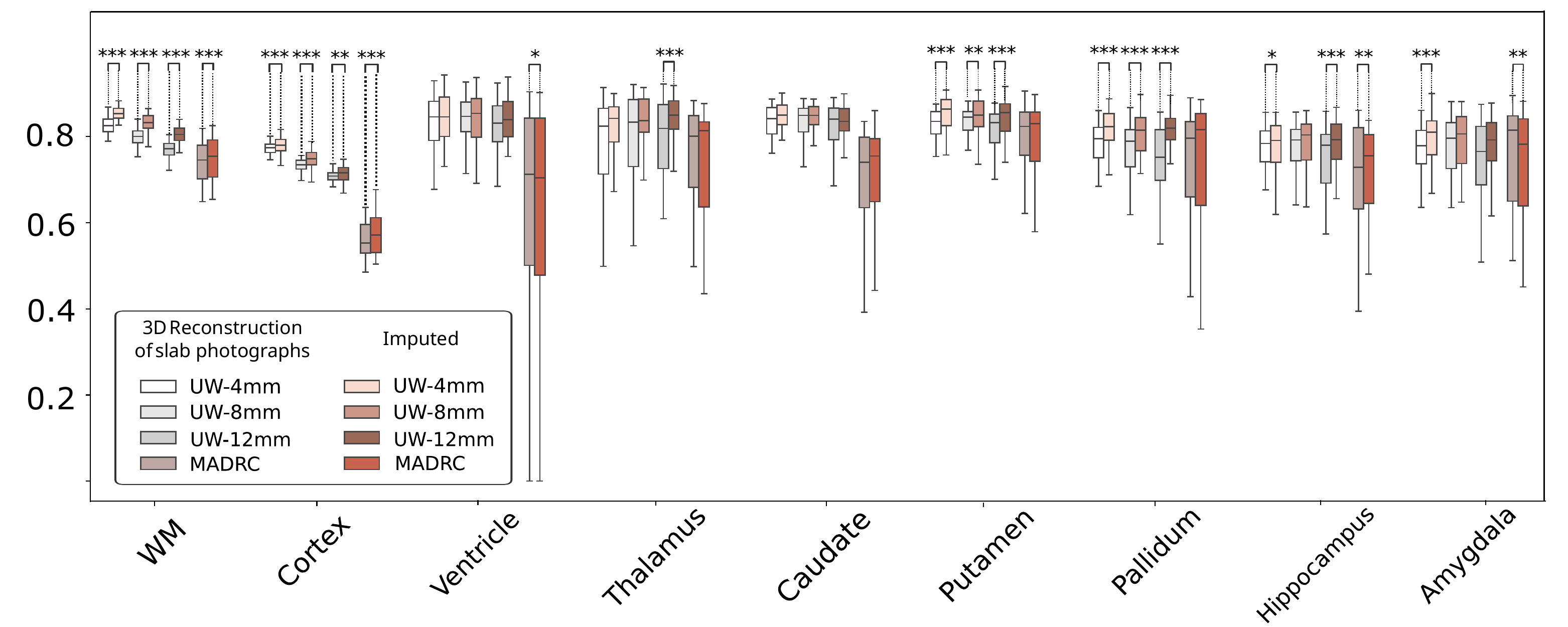}
    \caption{Box plots of region-specific Dice scores for automatic segmentations of 3D-reconstructed photo volumes, before and after imputations (gold standard is MRI). The marker $*$ indicates Wilcoxon p between $0.05$ and $0.01$, $**$ denotes $0.01\leq p <0.001$, and $***$ indicates $p <0.001$.}
    \label{fig:task_2_uwmadrc_dice}
\end{figure}

These findings are corroborated by the quantitative results shown in Figure~\ref{fig:task_2_uwmadrc_dice}. Region-specific Dice scores demonstrate the superior performance of our imputation method compared to the original photo reconstructions (Wilcoxon $p < 0.001$, Table S~\ref{app:task_2_dice_scores_pvalues}). 
Compared to the original 3D reconstructions of slab photographs, our proposed imputation method achieves greater Dice scores for almost all regions across both datasets, with the greatest improvement in Dice observed for the cortex and white matter, thanks to the improved cortical definition -- as reported in the previous section (Figure~\ref{fig:task_1_uw_surface_recon_example}). 
Improvements in subcortical regions are also observed across the board, with only two regions (out of 9) showing slight decreases in performance (ventricle and amygdala in MADRC).

\subsection{Atlas Registration} 
Finally, we evaluate the impact of our slice imputation on atlas registration, a downstream task particularly sensitive to anatomical consistency. 
Atlas-based registration can be used in digital guided dissection, to ensure that corresponding regions are sampled for different subjects, despite substantial inter-subject anatomical variability -- especially in cortical areas.
Beyond atlas registration, the aggregation of histopathological information from different subjects in a common coordinate frame enables group-wise analyses, and supports a wide range of downstream functional studies.


Here, we register the MNI-ICBM152 atlas to each photo-reconstruction using the NiftyReg registration tool \citep{modat2010fast}. 
This software enables non-rigid registrations of 2D and 3D images based on cubic B-splines. As such, equally spaced control points are defined over the reference images (here, the photo-reconstructions), and are mapped locally to the floating image (the atlas).
With the deformation fields estimated for each registration, the atlas segmentation map, computed with SynthSeg on the atlas prior to registration, is then resampled with nearest-neighbor interpolation. The resampled atlas segmentations are compared with gold-standard segmentations from the MRIs using Dice coefficients.

\begin{figure}[h!]
    \centering
    \includegraphics[width=\linewidth]{ 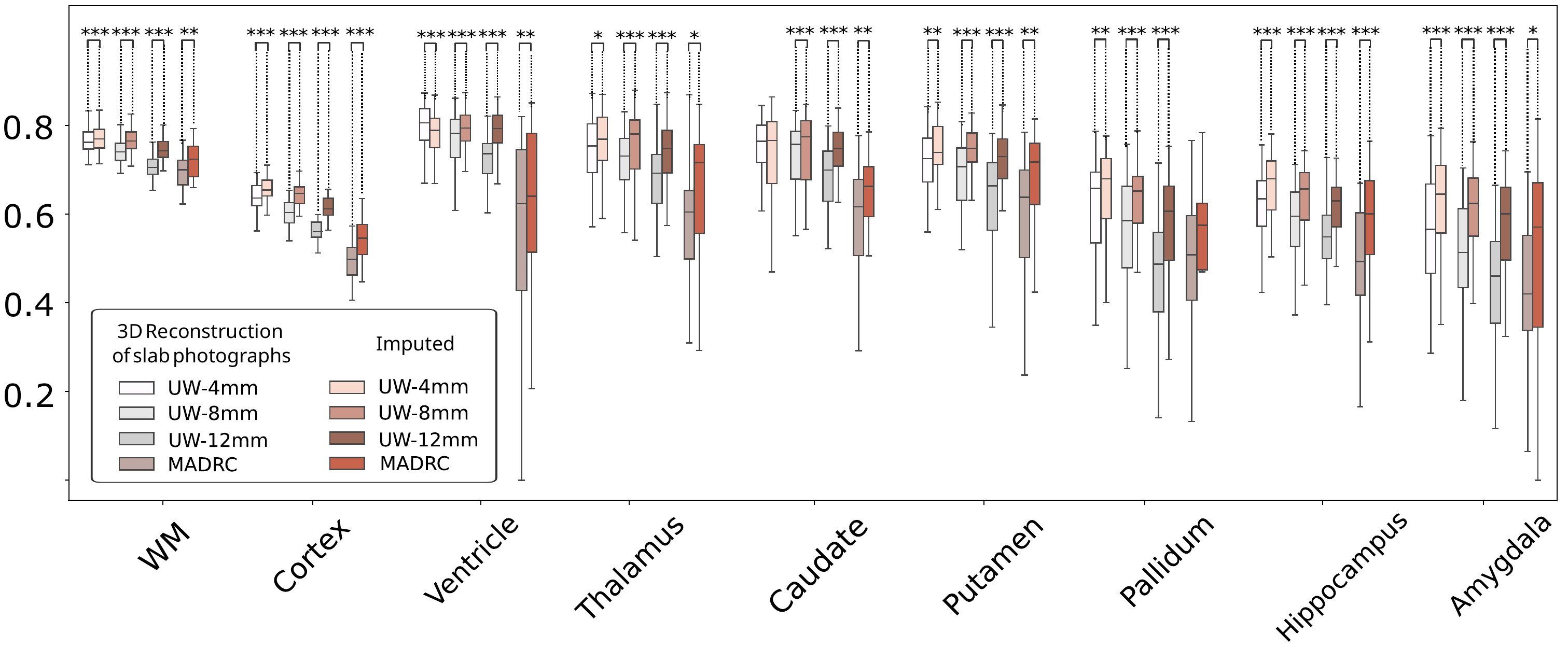}
    \caption{Box plots of Region-specific Dice scores of the (warped) atlas segmentation and the gold standard segmentations derived from the MRI scans, for both UW and MADRC datasets. The marker $*$ indicates Wilcoxon p between $0.05$ and $0.01$, $**$ indicates $0.01\leq p <0.001$, and $***$ denotes $p <0.001$.}
    \label{fig:task_3_uwmadrc_dice_scores}
\end{figure}

Figure~\ref{fig:task_3_uwmadrc_dice_scores} shows results for the overlap between the MRI and (warped) atlas segmentations, before and after imputation, for the MADRC and UW datasets. These results highlight, once again, the positive performance of our proposed method on downstream analyses. 
With imputation, we see improved Dice scores for every dataset, region, and thickness  (Wilcoxon $p < 0.05$, Table S~\ref{app:task_3_dice_scores_pvalues_final}) with only two exceptions (out of 36 combinations): caudate reconstructions at 4 mm in the UW dataset and pallidum reconstructions in the MADRC dataset (see box plots in Figure~\ref{fig:task_3_uwmadrc_dice_scores}).

\section{Discussion and Conclusion}

Our previous work proposed a framework to reconstruct 3D brain volumes from 2D coronal slab photographs of dissected human brain tissue. However, the anisotropic resolution of these reconstructions decreased the precision of downstream morphometry -- particularly for cortical analysis.
In this work, we introduce a computationally efficient super-resolution step that enables recovery of high-resolution isotropic brain volumes from 3D photograph reconstructions. Using measurements derived from isotropic MRI as gold standard, we have shown that these imputed volumes greatly increase the accuracy of downstream analyses,  including volumetric segmentation, cortical analysis and atlas registration.

In surface reconstruction, our method recovers smoother and more anatomically plausible surfaces, compared to the blocky and voxelated outputs obtained from the previous method. This improved level of representation fidelity  extends to subcortical structures: compared with the baseline reconstructions, higher Dice scores are achieved with our proposed imputation across slab thickness and datasets. In atlas registration,  imputation  yields improved region-specific overlaps with the gold standard.

Future work will focus on reducing the gap between training and test distributions. While the current domain randomization scheme yields excellent generalizability, it does not fully model some nuances of  of real images, e.g., light reflection, uneven thickness, traces of blood (for fresh tissue), or non-Gaussian intensity distributions. 
More accurate simulation of the slicing process or application of modern deep generative approaches may help reduce this gap. 

Imputation of slabs located near the brain boundaries could also be improved in future work. Compared to central slabs, which benefit from two informative input slices,  reconstruction  becomes significantly more difficult at the edges of the volume. In these regions, the model is only given one informative slice (and also, slices are typically thicker). As a result, imputation on the edges is qualitatively inferior compared with  central regions.

Furthermore, our method operates on the red, green, and blue channels independently, since pilot experiments on simulated RGB data increased computational requirements but did not yield any improvement. While this approach often works sufficiently well, it sometimes leads to chromatic artifacts. More advanced multi-channel simulations may be required to tackle this issue.  

Overall, this study demonstrates that introducing a lightweight super-resolution step into our 3D photograph reconstruction method substantially enhances the accuracy of subsequent morphometric analysis. Therefore, our improved approach provides an improved bridge between neuropathology and neuroimaging, bringing a new level of spatial precision to the study of morphological signatures of brain disease.


\section{Materials and Methods}

\subsection{Datasets}

We use three different datasets in this study: one for training and two for evaluation. The former comprises of 3D segmentations obtained from brain MRI scans; the latter two are datasets of real slab photographs.

\subsubsection{Training data}
MRI scans were gathered from 10 publicly available datasets: 
ABIDE \citep{di2014autism}, ADHD200 \citep{brown2012adhd},  ADNI  \citep{jack2008alzheimer_adni}, AIBL \citep{fowler2021fifteen_aibl},  COBRE \citep{mayer2013functional}, Chinese-HCP \citep{vogt2023chinese}, HCP \citep{van2012human_hcp}, ISBI2015 \citep{carass2017longitudinal_isbi}, MCIC \citep{gollub2013mcic}, and OASIS3 \citep{lamontagne2019oasis}.  
More specifically, we used 1~mm isotropic T1 scans, which we segmented with FreeSurfer \cite{fischl2012freesurfer}  and visually quality controlled (QCed), to yield a final training cohort with 5,279 cases.
In order to ensure generalizability at 
test time, we used a domain randomization approach to generate synthetic volumes  from the segmented MRI scans (details below). 

\subsubsection{Test data}

We leverage two datasets of \textit{postmortem} specimens from the Alzheimer's Disease Research Center at the Precision
Neuropathology Core, UW School of Medicine; and the MADRC. 

\noindent\textbf{UW}:
This dataset comprises dissection photography of fixed tissue and paired \textit{ex vivo} MRI scans for 28 specimens. After routine extraction and fixation in 10\% neutral buffered formalin, the cerebellum and brainstem were excised. The cerebrum was then embedded in agarose to minimize deformation and maintain tissue integrity during slicing. Specimens were scanned using a 3T MRI with a FLAIR sequence at near 1~mm isotropic resolution. The embedded specimens were slabbed at a consistent $\sim$4~mm thickness using a modified meat slicer. Slabs were photographed alongside perpendicular rulers, from the posterior side against a high contrast black background. 3D reconstructions were obtained using our registration strategy presented in \cite{gazula2024machine}. Gold-standard segmentations were obtained by segmenting the MRI scans with SynthSeg.

\noindent\textbf{MADRC}:
This dataset comprises photographs of coronal slabs from 19 left hemispheres and 9 cerebra. Post extraction, brain specimens were fully fixed in 10\% neutral buffered formalin and coronally slabbed by hand into slices with non-uniform thickness (approximately 8~mm thick, on average). The slabs were photographed against a black background alongside a ruler.
Surface scans were acquired using a turntable and an Einscan Pro HD scanner (Shining 3D, Hangzhou, China, 0.05 mm point accuracy), as in \cite{gazula2024machine}.
Companion MRI scans were downloaded from our hospital's Picture Archiving and Communication System (PACS). Since these were heterogeneous scans acquired for clinical purposes, we apply the following harmonization steps: \textit{(i)}~using SynthSR \citep{iglesias2023synthsr}, we obtain a synthetic 1~mm isotropic T1 volume from every scan in every MRI session; \textit{(ii)}~we then segment them with SynthSeg, which also produces automatic QC scores with an automated method \citep{billot2023robust}; and \textit{(iii)}~we keep the synthetic T1 and segmentation maps with the highest QC score; \textit{(iv)} we register the synthetic T1s to the 3D photograph reconstructions using NiftyReg \citep{modat2010fast}; and \textit{(v)}~the resulting deformation fields are used to warp the corresponding labels and obtain gold-standard segmentations in photograph space. 

The UW dataset is a high-quality sample that enables us to test the accuracy of our methods on real data as a function of slab thickness at 4~mm intervals. The MADRC dataset enables an evaluation of our methods on real images that are more representative of the data that are typically acquired in most brain banks.

\subsection{Methods}

\subsubsection{Data generator}
Our approach relies on synthetic data generated on-the-fly during training, using a two-step approach: 3D synthesis with domain randomization, and random digital sectioning of coronal slices. 
The 3D synthesis uses a domain randomization scheme that we have successfully used in prior work \citep{billot2023robust,iglesias2023synthsr,gazula2024machine,gopinath2025recon}, and which we summarize here.

Generation starts by randomly selecting one of the 1~mm isotropic segmentations from the training dataset. This is geometrically augmented with random linear and nonlinear transformations. As in \cite{gazula2024machine},  the nonlinear component varies more quickly along the anterior-posterior axis in order to model the imperfect 3D reconstruction of real photograph slabs encountered at test time.
A Gaussian Mixture Model (GMM) conditioned on the deformed isotropic segmentation, is then used to generate a synthetic, 1~mm isotropic image. The parameters of the GMM (means and variances) are randomized to make the network agnostic to variations in image appearance.
Next, we apply gamma intensity augmentation  to randomly skew the Gaussian distributions. The 3D synthesis ends with the random simulation of heterogeneous illumination. This is achieved via a smooth, multiplicative field that, as the nonlinear deformation, varies more quickly across the anterior-posterior direction, in order to model uneven illumination across coronal slabs.

The random digital slabbing amounts to selecting a random anterior-posterior coordinate and a random slab thickness $d$ between 2 and 12 mm. These are used to generate three coronal images: two images $d$~mm apart, which model the input photographs, and one at a random coordinate within the slab, which models the slice to impute.  We sample 32 triplets from the synthetic 3D volume at every iteration to form a minibatch.

\subsubsection{AI slice imputation}
We approach imputation as an image-to-image prediction method: given two parallel coronal slabs and a relative coordinate in between, the goal is to impute the photograph at the specified location. Rather than predicting the image intensities directly, we train a neural network to estimate the residual relative to a linear interpolation of the intensities. In practice, this approach converges much faster because the linear interpolation already provides a reasonable approximation to the solution. The linear interpolation is given by:
\begin{equation}\label{eq:linear_interpolation}
    \bm{y}_{lin}(\bm{x}_1, \bm{x}_2, d_1, d_2) = \frac{d_2}{d}\;\bm{x}_1 + \frac{d_1}{d}\;\bm{x}_2,
\end{equation}
where $\bm{x}_1$, and $\bm{x}_2$ are the images of the two slabs; $d_1$ and $d_2$ are the distances from the image to impute to $\bm{x}_1$ and $\bm{x}_2$, respectively; and $d= d_1+d_2$ is the total thickness. 

\begin{figure}[h!]
    \centering
    \includegraphics[width=0.85\linewidth]{ 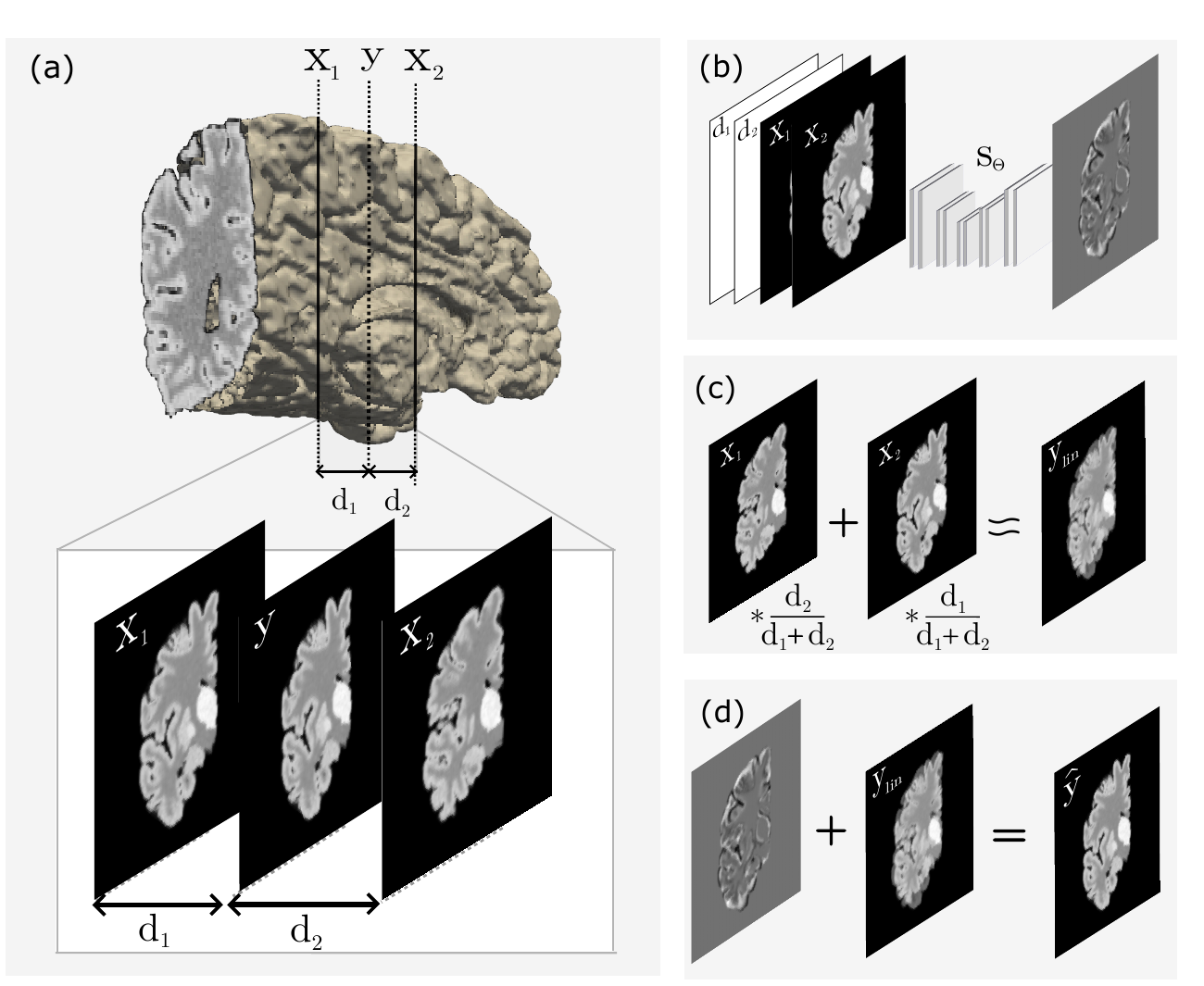}
    \caption{Machine learning interpolation framework. (a) Example of a training sample illustrating a slab of thickness ${d}_1 + {d}_2$ with input slices $\bm{x}_1$ and $\bm{x}_2$, and the intermediate target slice $\bm{y}$. (b) Residual prediction of the network $\bm{S}_\theta(x_1, x_2, d_1, d_2)$, conditioned on input slices ($\bm{x}_1$, $\bm{x}_2$), and the  respective distances to the target slice (${d}_1$, ${d}_2$).
    (c) Linear interpolation $\bm{y}_{lin}$ computed from the input slices ($\bm{x}_1$, $\bm{x}_2$) weighted according to the relative distances ${d}_1$, and ${d}_2$. (d) Final imputed slice $\bm{\hat{y}}$, obtained by adding the predicted residual to the linear interpolation $\bm{y}_{lin}$.}
    \label{fig:interpolation_framework}
\end{figure}

Our method imputes slices as:
\begin{equation}
      \hat{\bm{y}}(\bm{x}_1, \bm{x}_2, d_1, d_2) = \bm{y}_{lin}(\bm{x}_1, \bm{x}_2, d_1, d_2) + \bm{S}_{\theta}(\bm{x}_1, \bm{x}_2, d_1, d_2),
\end{equation}
where $\bm{S}_{\theta}$ is the prediction of a 2D U-Net \citep{ronneberger2015u} parameterized by $\theta$. In practice, $d_1$ and $d_2$ are replicated into 2D arrays of the same size as $\bm{x}_1,\bm{x}_2$ and concatenated with those into inputs with four channels. The framework is summarized in Figure~\ref{fig:interpolation_framework}. 

The U-Net comprises of four encoder and four decoder layers with symmetric blocks of downsampling and upsampling modules. 
Each encoder block contains two sequential units, each composed of Group Normalization, a 3×3 convolution, and a Leaky ReLU activation, followed by a 2×2 max-pooling layer. At each encoder stage, the spatial resolution is halved while the number of feature channels is doubled.
The bottleneck representation reaches a latent dimensionality of 1024 channels with the most compact and semantically rich description of the input. The decoder architecture mirrors the encoder, progressively restoring the spatial resolution with interpolation-based upsampling and skip connections implemented via feature concatenation.

The U-Net parameters $\theta$ were optimized by minimizing a loss combining: \textit{(i)}~the mean absolute error (MAE) between the (min-max normalized) pixels intensities of the predicted slice $\hat{\bm{y}}$ and the ground truth; and \textit{(ii)}~the MAE between their gradient magnitudes, computed with a Sobel operator. We trained the model until convergence, determined with the MAE on a validation set of $1,000$ synthetic slices. 
We optimized the training loss on batches of $32$ synthetic slabs, using the ``Adam'' optimizer with a learning rate of $10^{-6}$.

At test time, the trained U-Net is used to impute slices at fixed 1~mm intervals in the anterior-posterior direction, irrespective of the slice spacing of the provided 3D photograph reconstruction. Inference is performed on the red, green, and blue channels independently.

\section{Acknowledgment}
This research was primarily supported by the National Institute on Aging (R01AG070988). Additional support was provided by NIH grants RF1AG080371 and 1R21NS138995. 
The UW BioRepository and Integrated Neuropathology (BRaIN) Laboratory and Precision Neuropathology Core are supported by the National Institutes of Health (NIH) through the UW Alzheimer’s Disease Research Center (P30AG066509), the Adult Changes in Thought (ACT) study (U19AG066567 and R01AG060942), 
the BRAIN Initiative Cell Atlas Network (UM1MH130981 and UM1MH134812), the Seattle Alzheimer’s Disease Brain Cell Atlas (U19AG060909), the US Department of Defense (DoD W81XWH-21-S-TBIPH2), 
the Chan-Zuckerberg Initiative,
cooperative agreements (U24AG072458; U24NS133949; U24NS133945; U24NS135651; U01NS137500; and U01NS137484), 
 and the Allen Institute for Brain Science. 
Dr. Keene is additionally supported as a Weill Neurohub Investigator and the Nancy and Buster Alvord Endowed Chair in Neuropathology. 
We are deeply grateful to the research participants and their families without whom this work would be impossible.

\newpage

\bibliographystyle{unsrtnat} 
\bibliography{references}
\newpage

\setcounter{figure}{0}
\setcounter{table}{0}
\setcounter{section}{0}
\label{first:app}
\section{Supplement}

\subsection{Extended results of Surface reconstructions}

\begin{figure}[h!]
    \centering
    \includegraphics[width=0.7\linewidth]{ 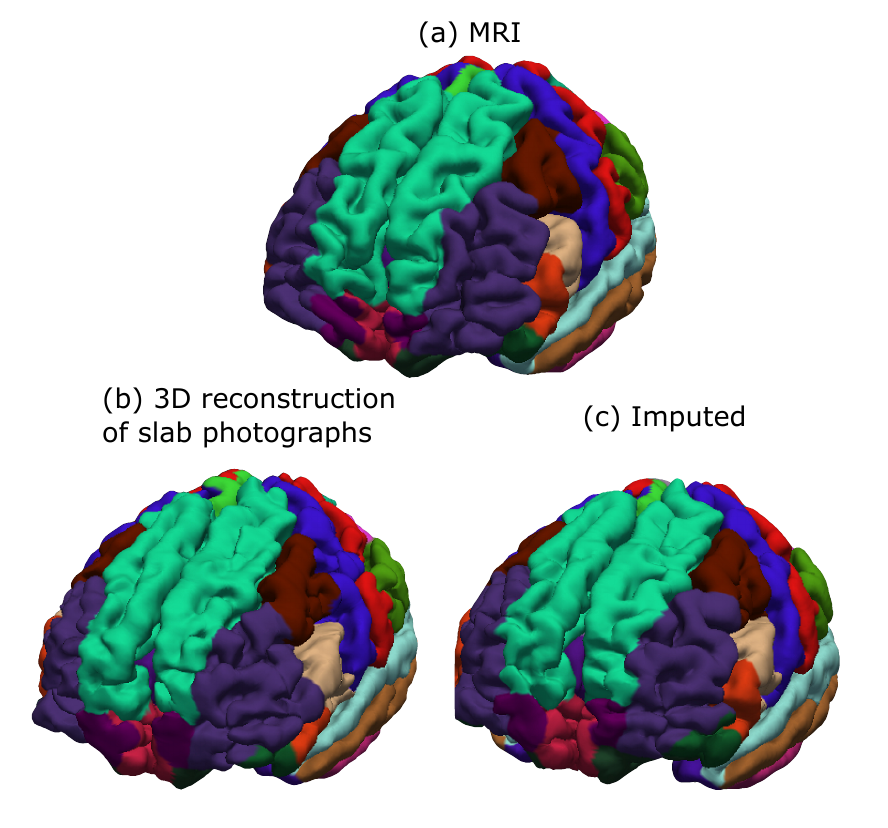}
    \caption{Pial surfaces with overlaid parcellations, computed with Recon-Any on one case from the MADRC dataset. (a)~Reference surface from MRI (gold standard). (b)~Surface obtained from the 3D reconstruction of slab photographs. (c)~Surface obtained with the proposed imputation.}
\label{app:task_1_madrc_illustration_surfaces}
\end{figure}

\begin{figure}[h!]
    \centering
    \includegraphics[width=.85\linewidth]{ 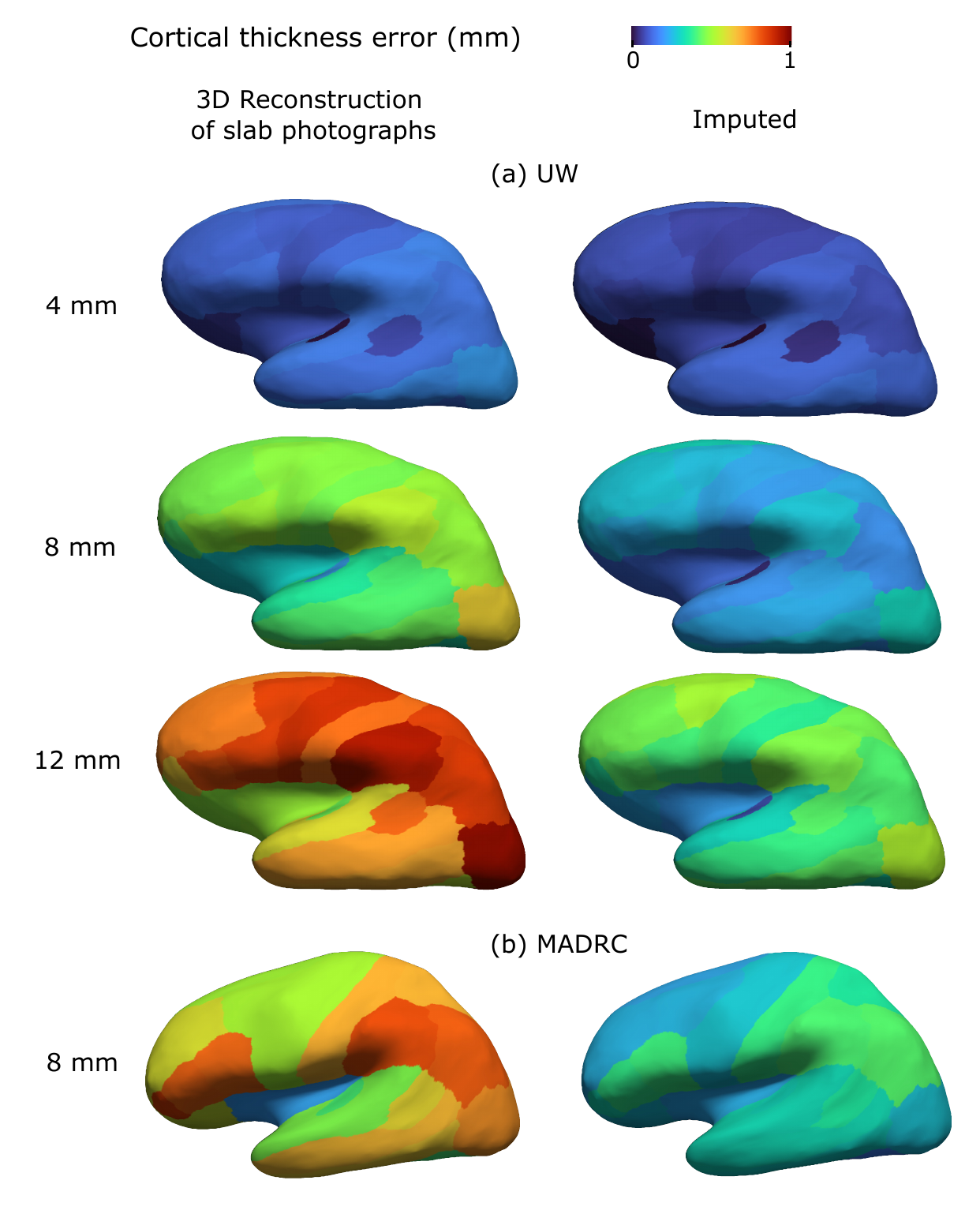}
    \caption{Color coded illustration of cortical thickness error distributions of the original (left) and our proposed method of imputation (right), overlaid on inflated surface hemispheres computed with Recon-Any. (a) Distribution of errors from reconstructions of UW dataset at three slab thicknesses (4, 8, 12 mm) (b) Distribution of errors from MADRC reconstructions (8 mm).}
\label{app:task_1_uw_madr_cortical_thickness_ilustration}
\end{figure}

\begin{table*}[h!]
\centering
\caption{Surface and thickness errors (in mm) for Recon-Any  of 3D photo reconstructions, computed against gold-standard MRI references. P-values from Wilcoxon Rank Sum statistical tests comparing both methods are reported for all evaluations.}
\begin{tabular}{lcccc}
\toprule
\multicolumn{5}{c}{\textbf{Pial Surface Error}} \\ 
\midrule
\textbf{Dataset} & \textbf{UW – 4 mm} & \textbf{UW – 8 mm} & \textbf{UW – 12 mm} & \textbf{MADRC} \\
\textbf{Photo-recon} & 0.845 & 1.084 & 1.313 & 1.767 \\
\textbf{Imputed}     & 0.807 & 0.912 & 1.045 & 1.399 \\
\textbf{p-Value}  & $<$0.001 & $<$0.001 & $<$0.001 & $<$0.001 \\
\midrule
\multicolumn{5}{c}{\textbf{White Matter Surface Error}} \\ 
\midrule
\textbf{Dataset} & \textbf{UW – 4 mm} & \textbf{UW – 8 mm} & \textbf{UW – 12 mm} & \textbf{MADRC} \\
\textbf{Photo-recon} & 0.781 & 1.015 & 1.211 & 1.657 \\
\textbf{Imputed}     & 0.763 & 0.854 & 0.969 & 1.279 \\
\textbf{p-Value}  & $<$0.001 & $<$0.001 & $<$0.001 & $<$0.001 \\
\midrule
\multicolumn{5}{c}{\textbf{Cortical Thickness Error}} \\ 
\midrule
\textbf{Dataset} & \textbf{UW – 4 mm} & \textbf{UW – 8 mm} & \textbf{UW – 12 mm} & \textbf{MADRC} \\
\textbf{Photo-recon} & 0.330 & 0.460 & 0.552 & 0.704 \\
\textbf{Imputed}     & 0.261 & 0.276 & 0.331 & 0.367 \\
\textbf{p-Value}  & $<$0.001 & $<$0.001 & $<$0.001 & $<$0.001 \\
\bottomrule
\end{tabular}
\label{app:surface_cortical_errors_updated}
\end{table*}

\clearpage

\subsection{Extended results of Volume segmentations}
\begin{table*}[h!]
\centering
\caption{Region-specific Dice scores of automated segmentations of 3D reconstructions of photographs, before and after imputation. The gold standard segmentatations are obtained from MRI scans. The p-values are from Wilcoxon Rank tests (non-parametric, paired).}
\begin{tabular}{lccc}
\toprule
\multicolumn{4}{c}{\textbf{MADRC}} \\ 
\midrule
\textbf{Region} & \textbf{Photo-recon} & \textbf{Imputed} & \textbf{p-Value} \\
Amygdala    & 0.727 & 0.695 & 0.001 \\
Caudate     & 0.689 & 0.690 & 0.837 \\
Cortex      & 0.505 & 0.522 & $<$0.001 \\
Hippocampus & 0.716 & 0.698 & $<$0.001 \\
Pallidum    & 0.712 & 0.716 & 0.554 \\
Putamen     & 0.772 & 0.774 & 0.736 \\
Thalamus    & 0.722 & 0.719 & 0.388 \\
Ventricle   & 0.631 & 0.624 & 0.003 \\
WM          & 0.670 & 0.678 & $<$0.001 \\
\midrule
\multicolumn{4}{c}{\textbf{UW – 4 mm}} \\ 
\midrule
Amygdala    & 0.698 & 0.751 & $<$0.001 \\
Caudate     & 0.825 & 0.815 & 0.017 \\
Cortex      & 0.770 & 0.778 & $<$0.001 \\
Hippocampus & 0.710 & 0.732 & 0.003 \\
Pallidum    & 0.714 & 0.793 & $<$0.001 \\
Putamen     & 0.797 & 0.845 & $<$0.001 \\
Thalamus    & 0.729 & 0.796 & 0.013 \\
Ventricle   & 0.824 & 0.831 & 0.091 \\
WM          & 0.825 & 0.850 & $<$0.001 \\
\midrule
\multicolumn{4}{c}{\textbf{UW – 8 mm}} \\ 
\midrule
Amygdala    & 0.720 & 0.754 & 0.060 \\
Caudate     & 0.825 & 0.827 & 0.023 \\
Cortex      & 0.733 & 0.747 & $<$0.001 \\
Hippocampus & 0.724 & 0.751 & 0.012 \\
Pallidum    & 0.731 & 0.786 & $<$0.001 \\
Putamen     & 0.815 & 0.836 & $<$0.001 \\
Thalamus    & 0.767 & 0.832 & 0.035 \\
Ventricle   & 0.831 & 0.838 & 0.336 \\
WM          & 0.798 & 0.830 & $<$0.001 \\
\midrule
\multicolumn{4}{c}{\textbf{UW – 12 mm}} \\ 
\midrule
Amygdala    & 0.722 & 0.739 & 0.033 \\
Caudate     & 0.817 & 0.819 & 0.273 \\
Cortex      & 0.707 & 0.712 & 0.001 \\
Hippocampus & 0.702 & 0.747 & $<$0.001 \\
Pallidum    & 0.694 & 0.799 & $<$0.001 \\
Putamen     & 0.802 & 0.837 & $<$0.001 \\
Thalamus    & 0.759 & 0.843 & $<$0.001 \\
Ventricle   & 0.820 & 0.829 & 0.028 \\
WM          & 0.768 & 0.803 & $<$0.001 \\
\bottomrule
\end{tabular}
\label{app:task_2_dice_scores_pvalues}
\end{table*}

\begin{figure}[h!]
    \centering
    \includegraphics[width=\linewidth]{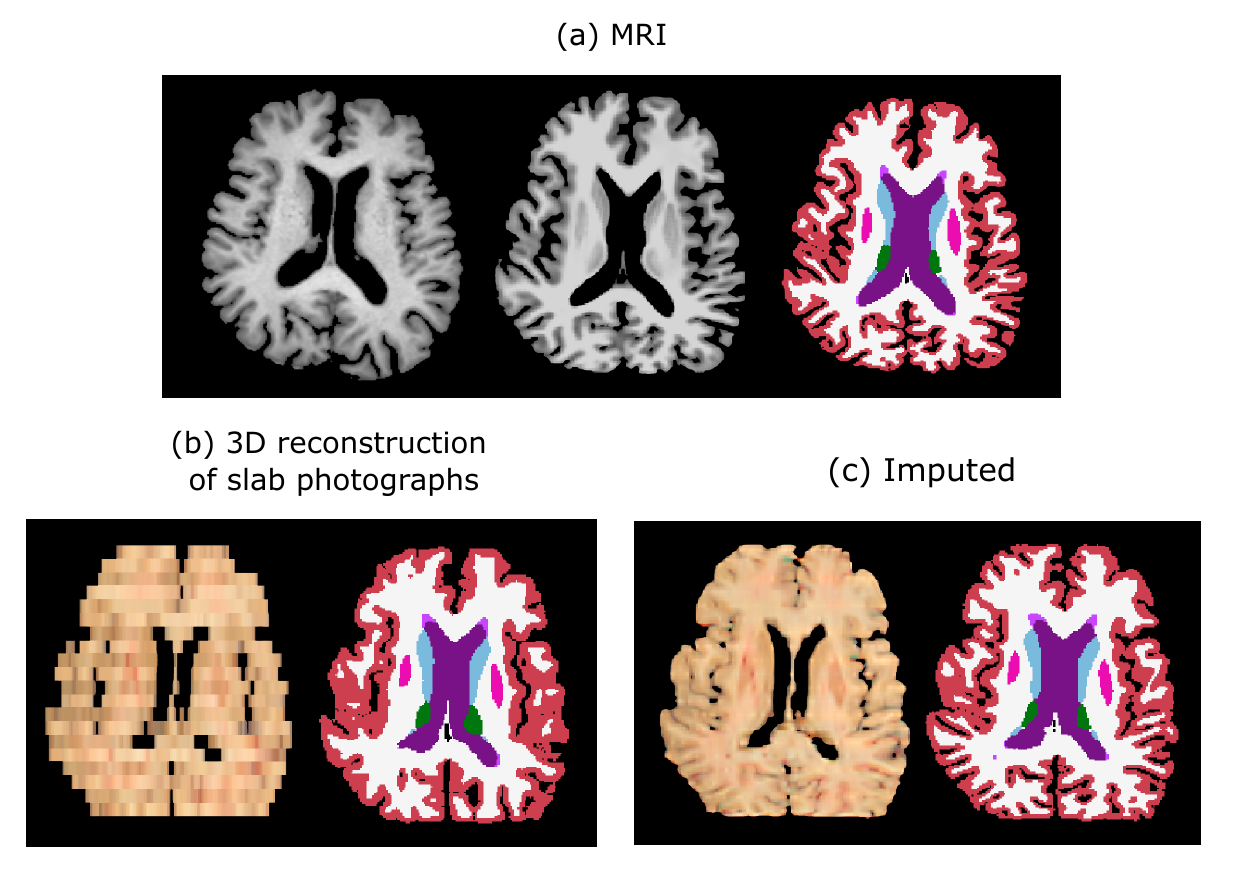}
    \caption{Axial view of automated segmentations from one example of the MADRC dataset. (a)~Native MRI prior to any pre-processing (left), 
    gold-standard MRI processed with ''SynthSR'' (middle), and automated segmentation obtained with ''SynthSeg'' (right). (b)~3D Reconstruction of slab photographs (left) and automated segmentation with ''Photo-SynthSeg'' (right). (c)~Imputed reconstruction (left) and automated segmentation with ''SynthSeg'' (right).}
\label{app:photosynthseg_segmentations_madrc}
\end{figure}

\clearpage
\subsection{Extended results of Atlas registrations}

\begin{table*}[h!]
\centering
\caption{Region-specific Dice scores of (warped) atlas segmentations and gold-standard segmentations (from the MRIs). P-values computed with a (paired, non-parametric) Wilcoxon Rank Sum tests are also reported across datasets and slab thicknesses.}
\begin{tabular}{lccc}
\toprule
\multicolumn{4}{c}{\textbf{MADRC}} \\ 
\midrule
\textbf{Region} & \textbf{Photo-recon} & \textbf{Imputed} & \textbf{p-Value} \\
Amygdala    & 0.409 & 0.503 & 0.005 \\
Caudate     & 0.535 & 0.613 & 0.001 \\
Cortex      & 0.449 & 0.493 & $<$0.001 \\
Hippocampus & 0.480 & 0.574 & $<$0.001 \\
Pallidum    & 0.450 & 0.497 & 0.056 \\
Putamen     & 0.574 & 0.646 & 0.001 \\
Thalamus    & 0.555 & 0.624 & 0.002 \\
Ventricle   & 0.545 & 0.581 & $<$0.001 \\
WM          & 0.629 & 0.656 & $<$0.001 \\
\midrule
\multicolumn{4}{c}{\textbf{UW – 4 mm}} \\ 
\midrule
Amygdala    & 0.539 & 0.605 & $<$0.001 \\
Caudate     & 0.752 & 0.735 & 0.099 \\
Cortex      & 0.639 & 0.658 & $<$0.001 \\
Hippocampus & 0.610 & 0.653 & $<$0.001 \\
Pallidum    & 0.604 & 0.625 & 0.001 \\
Putamen     & 0.716 & 0.743 & $<$0.001 \\
Thalamus    & 0.742 & 0.757 & 0.001 \\
Ventricle   & 0.796 & 0.778 & $<$0.001 \\
WM          & 0.764 & 0.769 & $<$0.001 \\
\midrule
\multicolumn{4}{c}{\textbf{UW – 8 mm}} \\ 
\midrule
Amygdala    & 0.485 & 0.592 & $<$0.001 \\
Caudate     & 0.726 & 0.749 & $<$0.001 \\
Cortex      & 0.602 & 0.643 & $<$0.001 \\
Hippocampus & 0.573 & 0.626 & $<$0.001 \\
Pallidum    & 0.561 & 0.617 & $<$0.001 \\
Putamen     & 0.689 & 0.736 & $<$0.001 \\
Thalamus    & 0.709 & 0.751 & $<$0.001 \\
Ventricle   & 0.764 & 0.785 & $<$0.001 \\
WM          & 0.738 & 0.764 & $<$0.001 \\
\midrule
\multicolumn{4}{c}{\textbf{UW – 12 mm}} \\ 
\midrule
Amygdala    & 0.436 & 0.568 & $<$0.001 \\
Caudate     & 0.676 & 0.739 & $<$0.001 \\
Cortex      & 0.562 & 0.613 & $<$0.001 \\
Hippocampus & 0.536 & 0.598 & $<$0.001 \\
Pallidum    & 0.446 & 0.550 & $<$0.001 \\
Putamen     & 0.632 & 0.710 & $<$0.001 \\
Thalamus    & 0.670 & 0.734 & $<$0.001 \\
Ventricle   & 0.720 & 0.782 & $<$0.001 \\
WM          & 0.707 & 0.745 & $<$0.001 \\
\bottomrule
\end{tabular}
\label{app:task_3_dice_scores_pvalues_final}
\end{table*}

\end{document}